\documentclass[10pt,twocolumn,letterpaper]{article}

\usepackage{cvpr}
\usepackage{times}
\usepackage{epsfig}
\usepackage{graphicx}
\usepackage{amsmath,amssymb}
\usepackage{bbm}
\usepackage{multirow}
\usepackage{caption}
\usepackage{algorithm}
\usepackage{algpseudocode}
\usepackage{dsfont}
\usepackage{cite}

\usepackage[breaklinks=true,bookmarks=false]{hyperref}

\cvprfinalcopy 

\def\cvprPaperID{****} 

\begin{document}

\title{Feature Boosting Network For 3D Pose Estimation}

\author{Jun~Liu$^{1}$\\{\tt\small jliu029@ntu.edu.sg}
   \and Henghui~Ding$^{1}$\\{\tt\small ding0093@ntu.edu.sg}
   \and Amir Shahroudy$^{2}$\\{\tt\small amirsh@chalmers.se}
	\and Ling-Yu~Duan$^{3}$\\{\tt\small lingyu@pku.edu.cn}
   \and Xudong~Jiang$^{1}$\\{\tt\small exdjiang@ntu.edu.sg}
   \and Gang~Wang$^{4}$\\{\tt\small gangwang6@gmail.com}
   \and Alex~C.~Kot$^{1}$\\{\tt\small eackot@ntu.edu.sg}
	\and $^1$ School of Electrical and Electronic Engineering,\\Nanyang Technological University, Singapore
   \\$^2$ Chalmers University of Technology, Sweden
   \\$^3$ Peking University, China \hspace{20pt} 
   \\$^4$ Alibaba Group, China
}

\maketitle

\begin{abstract}
In this paper, a feature boosting network is proposed for estimating 3D hand pose and 3D body pose from a single RGB image.
In this method, the features learned by the convolutional layers
are boosted with a new long short-term dependence-aware (LSTD) module,
which enables the intermediate convolutional feature maps
to perceive the graphical long short-term dependency
among different hand (or body) parts using the designed Graphical ConvLSTM.
Learning a set of features that are reliable and discriminatively representative of the pose of a hand (or body) part is difficult
due to the ambiguities, texture and illumination variation, and self-occlusion in the real application of 3D pose estimation.
To improve the reliability of the features for representing each body part and enhance the LSTD module,
we further introduce a context consistency gate (CCG) in this paper,
with which the convolutional feature maps are modulated according to their consistency with the context representations.
We evaluate the proposed method on challenging benchmark datasets for 3D hand pose estimation and 3D full body pose estimation.
Experimental results show the effectiveness of our method that achieves state-of-the-art performance on both of the tasks.
\end{abstract}

\section{Introduction}

3D pose estimation (estimating the locations of the joints of the human hand or body in 3D space) is a challenging and fast-growing research area, thanks to its wide applications
in gesture recognition, activity understanding, human-machine interaction, etc. \cite{andriluka2010monocular}.
Most of the existing works make use of highly constrained configurations \cite{chen20173d},
such as multi-view systems \cite{hofmann2012multi} and depth sensors \cite{shotton2011real}, to infer the 3D poses.
In this paper, we address the problem of 3D pose estimation from a single RGB image that is much easier to be captured in uncontrolled environments \cite{pavlakos2017coarse,tome2017lifting,nie2017monocular,zimmermann2017learning}.
This task is challenging due to the ambiguities in recovering the 3D information from a single 2D image,
the complex articulations and frequent occlusions of the hand (or body) parts,
and the large variation of clothing textures, camera viewpoints, and lighting conditions, etc.

Convolutional neural networks (CNNs) demonstrate their superior performance in various machine vision tasks, such as image classification and video analysis \cite{simonyan2014two}.
Recently, they have also been successfully applied to 3D pose estimation \cite{pavlakos2017coarse,zimmermann2017learning,zhou2017towards,sun2017compositional,tome2017lifting,luvizon20182d,zanfir2018monocular}.
In this paper, we construct our framework based on a CNN architecture.

Previous work on 3D pose estimation has shown the benefits of using the connection information of the body parts
to refine the pose estimation results or lift 2D pose to 3D space \cite{nie2017monocular}.
In this paper, we incorporate the complex dependency and correlation information among different parts to the convolutional features that contain very rich and representative information.
Specifically, a novel long short-term dependence-aware (LSTD) module is proposed,
which is embedded inside the CNN architecture to boost the intermediate convolutional feature maps for 3D pose estimation.

Our LSTD module is constructed based on the designed graphical convolutional long short-term memory (Graphical ConvLSTM).
In the image of a human hand (or body), 
there are complex dependency patterns among different parts.
Some joints are physically connected and obviously correlated,
while some others can have indirect correlation in their motion and appearance.
In order to utilize these complex dependency patterns effectively,
we design a Graphical ConvLSTM for the LSTD module,
which enables the feature maps of each part to learn the longer-term (indirect) and shorter-term (direct) dependency relations to other parts.
By modeling the graphical long short-term dependency information among the features of different hand (or body) parts,
the boosted features produced by our LSTD module are very effective for 3D pose estimation.

The inputs of the proposed LSTD module for feature boosting are convnet feature maps
that represent the information for each hand (or body) part.
However, these feature maps which are extracted by the convolutional layers from a single 2D image, may be unreliable for representing the corresponding part,
due to the existence of ambiguities in 3D pose estimation, the frequent occlusions, and also the texture and lighting condition variations.
In order to mitigate this drawback,
we further improve the design of the LSTD module by adding a soft modulator, context consistency gate (CCG),
which assesses the consistency of the convolutional features with their context information
and modulates these features accordingly for boosting.

In our method, multiple convolutional layers and LSTD modules can be stacked sequentially to construct a deep feature boosting network.
In the whole convolutional architecture, the intermediate feature maps are boosted
at multiple levels of the network. 

The main contributions of this paper are summarized as follows:
  (1) We propose an LSTD module within the CNN architecture to boost the convolutional feature maps
  by allowing them to perceive the graphical long short-term dependency
  with the designed Graphical ConvLSTM.
  (2) We further improve the design of the LSTD module by adding a gating mechanism, CCG,
  to analyze the context consistency of the convolutional feature maps.
  The CCG acts as a soft modulator to regulate the propagation of the feature map information
  based on their context consistency,
  which also gives the LSTD module better insight about how to boost the feature maps.
  (3) The proposed end-to-end feature boosting network achieves state-of-the-art performance on challenging datasets for 3D hand pose estimation and 3D full body pose estimation.

The rest of this paper is organized as follows.
The related works are introduced in section \ref{sec:relatedwork}.
The proposed feature boosting network is described in detail in section \ref{sec:method}.
The experimental results are provided in section \ref{sec:experiments}.
Finally, we conclude the paper in section \ref{sec:conclusion}.






\section{Related Work}
\label{sec:relatedwork}

\subsection{3D Pose Estimation}
Different aspects of human hand (and body) pose estimation have been explored in the past few years \cite{sarafianos20163d,erol2007vision}.
We limit our review to more recent CNN-based approaches for 3D pose estimation.
These methods mainly fall into two categories: 3D regression-based, and intermediate 2D pose-based methods \cite{tome2017lifting}.

{\bf 3D regression-based methods:}
Many previous methods directly regress the 3D locations of each joint using the convolutional features.
For example, Li and Chan \cite{li20143d} designed a pretraining strategy, in which the 3D pose regressor was initialized with a model trained for body part detection.
Tekin \etal \cite{tekin2016structured} used auto-encoders to learn structured latent representations for 3D pose regression from the images.
Park \etal \cite{park20163d} introduced a CNN framework by simultaneously training for both 2D joint classification and 3D joint regression.
Ghezelghieh \etal \cite{ghezelghieh2016learning} proposed to learn the camera viewpoint based on CNNs to improve the performance of 3D body pose estimation.

{\bf Intermediate 2D pose-based methods:}
A very recent trend of works started to investigate a pipeline framework to strengthen the estimation of 3D poses.
In this pipeline framework, heatmaps of the joints are estimated in the 2D frames first.
These 2D poses are then regarded as the intermediate representations, and the 3D poses are estimated based on them.
For example,
Chen \etal \cite{chen20173d} combined the 2D pose estimation results and a 3D matching library,
and achieved promising performance for 3D human pose estimation.
Zimmermann \etal \cite{zimmermann2017learning} adopted a PoseNet to infer the 2D hand joint locations, and then used a PosePrior network to estimate the most likely 3D structure of the hand.
Zhou \etal \cite{zhou2017towards} augmented the 2D pose estimation sub-network with a 3D depth regression sub-network to perform 3D human pose estimation.
Tome \etal \cite{tome2017lifting} proposed to perform 2D joint estimation and 3D pose reconstruction jointly to improve both tasks.
Nie \etal \cite{nie2017monocular} proposed to predict the depth of joints based on the 2D joint locations and the body part image features for 3D pose estimation.

Our proposed method is based on the pipeline framework as mentioned above \cite{zhou2017towards,zimmermann2017learning},
\ie, the intermediate 2D poses are estimated for the final 3D pose estimation.
Different from these works on 3D pose estimation,
in our method,
the feature maps within the convolutional network are boosted by enabling them to perceive the long short-term dependency patterns among different parts with the proposed LSTD module.
Besides, a soft modulator, CCG, is added to analyze the reliability and context consistency of the convolutional features,
which encourages the network to learn reliable features for 3D pose estimation.

\subsection{Dependency Structure}
The analysis of the correlation between parts of the hand (or body) has been shown to be very useful for pose estimation.
Felzenszwalb \etal \cite{felzenszwalb2005pictorial} proposed to represent the human body by a collection of parts arranged in a deformable configuration for pose estimation.
Yang \etal \cite{yang2013articulated} described a method for articulated human detection and pose estimation in static images
based on the representation of deformable part models with a tree structure.
Chu \etal \cite{chu2016structured} introduced a structured feature learning method to reason the relationships within the body joints for 2D pose estimation.
Chen \etal \cite{chen2014articulated} proposed a graphical model of the body joints as a post-processing step. 

Different from the above-mentioned works,
in this paper, we propose a new LSTD module with Graphical ConvLSTM for feature boosting.
By introducing the Graphical ConvLSTM, we add an extra layer of feature analysis, to model the graphical long short-term dependency relations among different parts.
We show that the boosted feature maps derived from the LSTD module are more powerful for 3D pose estimation than the features before boosting.
Specifically, LSTD modules can be added at different layers,
thus the features in the whole CNN architecture can be boosted layer by layer.
Moreover, we introduce a gating mechanism (CCG) to flexibly regulate the propagation of the intermediate feature representations within the CNN architecture by analyzing their reliability and context consistency. 


\subsection{Gating Mechanism}

\begin{figure*}[!t]
\centering
   \includegraphics[scale=0.52,trim={8.5cm 5.0cm 6.8cm 4.399cm},clip]{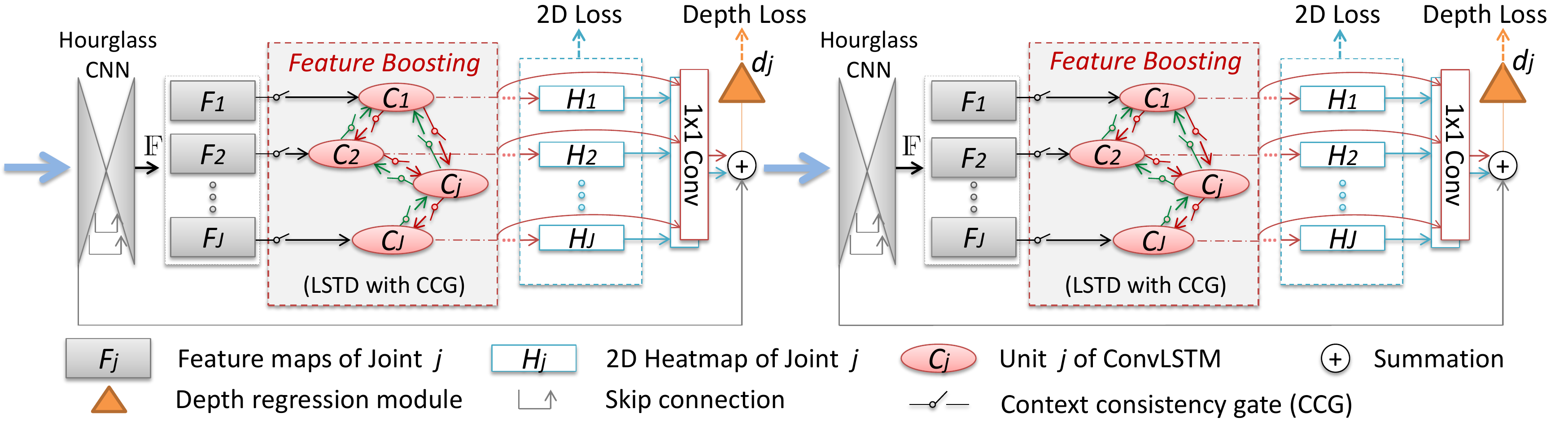}
\caption{
Illustration of the feature boosting network for 3D pose estimation.
The whole network is stacked with multiple similar sub-networks
(two sub-networks are used in our implementation).
The input of the first sub-network is an RGB image of a human hand (or a full human body).
The inputs of the latter sub-network are the concatenated feature maps from its previous sub-network.
In each sub-network, the Hourglass CNN layers \cite{newell2016stacked} are used to learn the convolutional features,
then the feature maps for different joints are fed to the LSTD module with CCG for feature boosting.
The boosted feature maps of each joint ($j$) are fed to the subsequent CNN layers to generate the 2D heatmap ($H_j$). 
Depth information ($d_j$) of each joint is regressed based on the summation of the boosted feature maps and the 2D heatmap representations 
(the feature maps obtained by this summation are also concatenated and fed to the subsequent sub-network as input for further feature boosting).
}
\label{Figure:Network_Architecture}
\end{figure*}

Our proposed context consistency gate (CCG) is inspired by the gating mechanism \cite{hochreiter1997long,varior2016gated,cho2014learning,liu2016eccv,xiong2016dynamic},
which is shown to be an important technique to improve the representation strength of deep networks.
Cho \etal \cite{cho2014learning} proposed a network with gated units to modulate the information flow for machine translation.
Xiong \etal \cite{xiong2016dynamic} designed an attention gate to explore the important information for textual and visual question answering.
Liu \etal \cite{liu2016eccv} introduced a trust gating mechanism to deal with the noisy sequences for activity analysis.
Dauphin \etal \cite{dauphin2017language} proposed gated linear units within the deep network for language modeling.

Compared to the aforementioned methods,
our soft modulator, CCG, is designed in a different context in terms of both its purpose and architecture.
The goal of the CCG is to assess the reliability of the convolutional features,
and accordingly regulate the propagation of them in the CNN architecture.
To the best of our knowledge, the proposed work is the first of its nature
in introducing gating mechanisms \cite{liu2016eccv} in a CNN architecture
for modulating and propagating the features by considering their context consistency for 3D pose estimation.


\section{The Proposed Method}
\label{sec:method}

Given a single RGB image of a human hand (or a full human body),
our goal is to estimate the locations of the major joints of the hand (or body) in 3D space.
In this paper, we propose a feature boosting network based on a CNN framework for this task.
A long short-term dependence-aware (LSTD) module is proposed,
which is embedded inside the CNN framework,
to boost the convolutional features
by enabling them to perceive the graphical long short-term dependency patterns among different parts.
Moreover, the design of the LSTD module is further improved by adding a context consistency gate (CCG),
which acts as a soft modulator to adjust the propagation of features through the network,
according to the context consistency and reliability.
The overall architecture of the feature boosting network is illustrated in \figurename{~\ref{Figure:Network_Architecture}}.

\subsection{Long Short-Term Dependence-aware Module}
\label{sec:method:convlstm}

\begin{figure}
	\begin{minipage}[b]{0.49\linewidth}
		\centering
		\centerline{\includegraphics[scale=0.2,trim={11.8cm 0.6cm 18.8cm 0.6cm},clip]{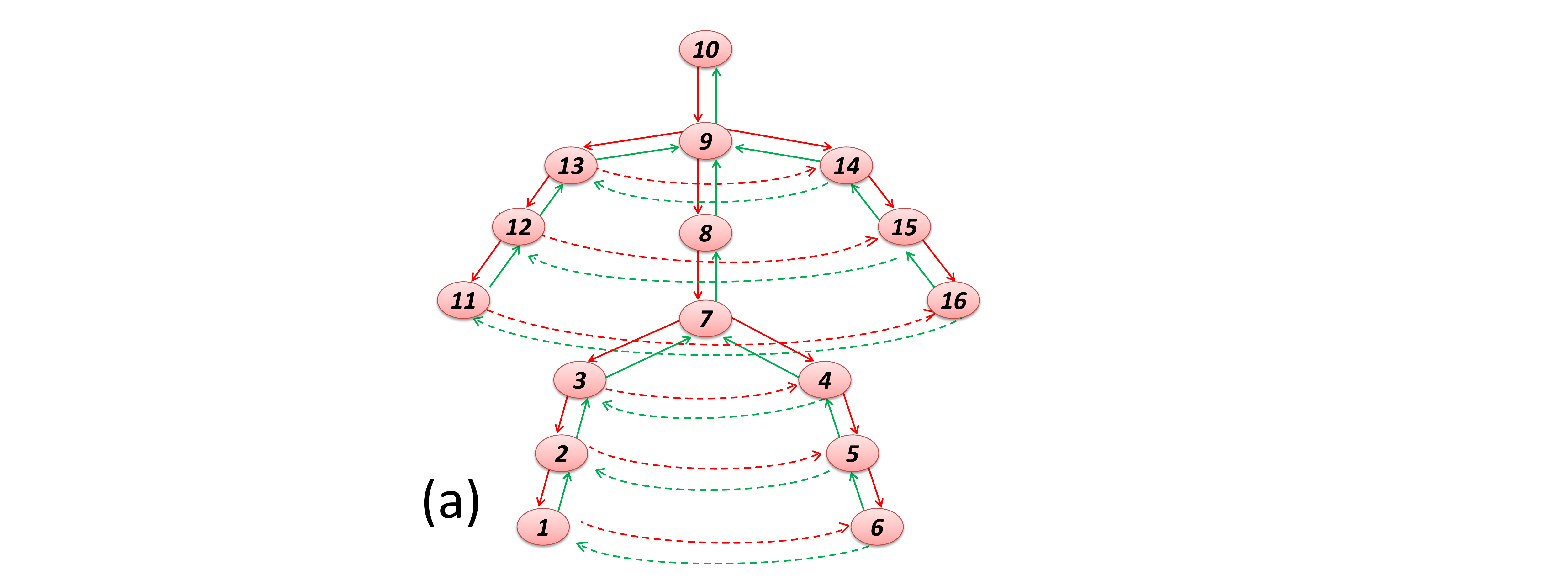}}
	\end{minipage}
	\begin{minipage}[b]{0.49\linewidth}
		\centering
		\centerline{\includegraphics[scale=0.22,trim={11.8cm 2.6cm 18.8cm 0.6cm},clip]{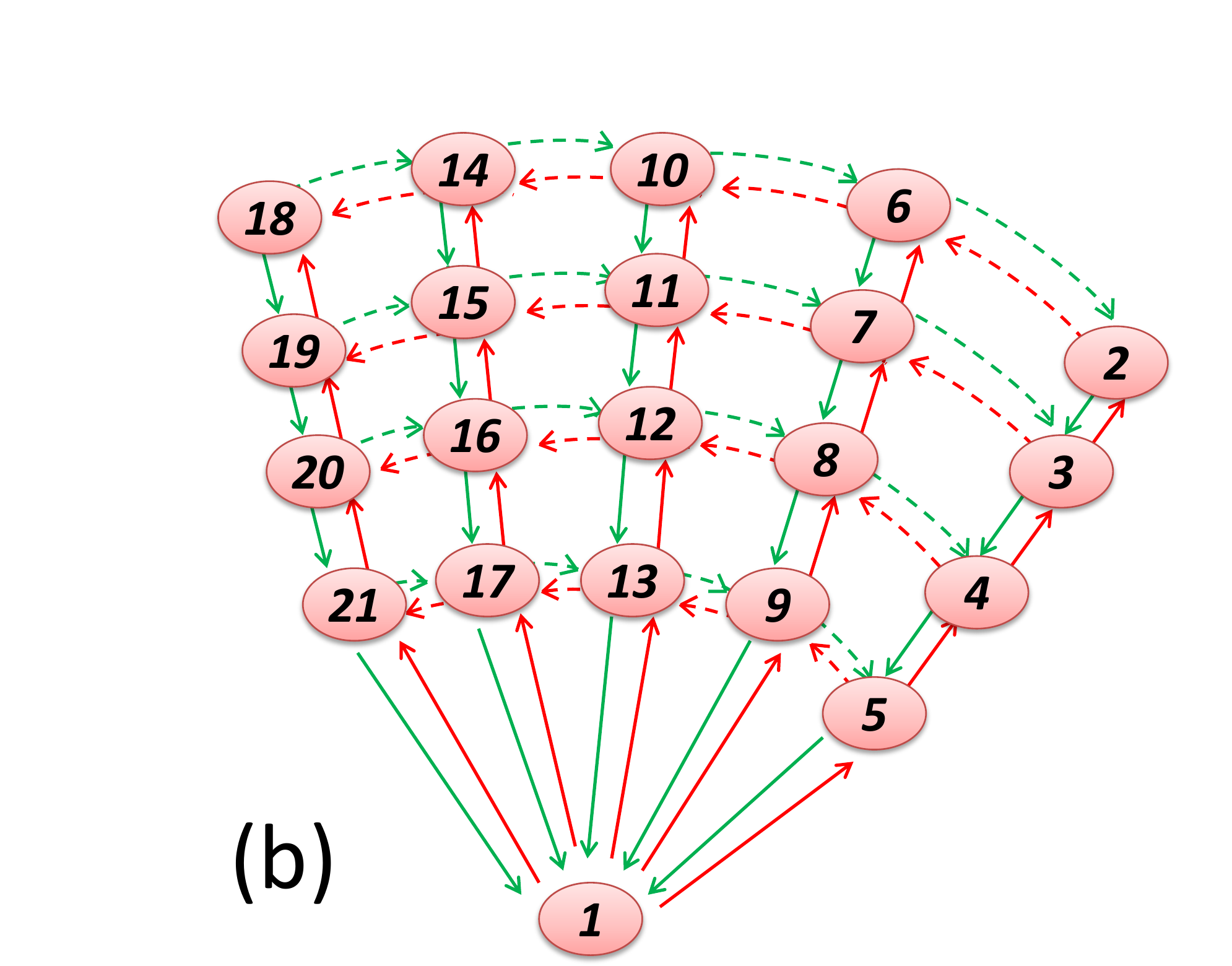}}
	\end{minipage}
	\caption{
Graphical long short-term dependency relationship between different parts (joints) of (a) full human body, and (b) human hand.
Solid lines denote the physical connections.
Dashed lines indicate the ``symmetrical'' relations.
}
	\label{fig:hand_human_lstm}
\end{figure}

There are direct and indirect kinematic dependency relations among different parts of the human hand (or body).
For example,
in \figurename{~\ref{fig:hand_human_lstm}(a)},
the adjacent joints, 2 and 3, are directly connected in the human body,
while the joints 2 and 7 are indirectly connected.
Utilizing these complex direct and indirect dependency patterns as a feature analysis step is beneficial for 3D pose estimation.

Many existing CNN-based 3D pose estimation approaches do not explicitly use the dependency structure,
while some others often consider it at the result level,
\eg, employ the dependency relations to refine the 3D estimations,
or use them to lift the 2D coordinates of the joints to 3D space at a post-processing stage \cite{nie2017monocular}.
In this paper,
we employ the direct and indirect dependency patterns to boost the intermediate features at different levels of the convolutional architecture for 3D pose estimation.
Specifically,
we introduce a novel long short-term dependence-aware (LSTD) module
to enable the features of each part of a hand (or a body)
to discover its long short-term dependency relations to other parts. 
Below we introduce the mechanism of the proposed LSTD module in detail.

\textbf{Graphical Dependency Relations.}
The major joints of the human body and hand
are illustrated in \figurename{~\ref{fig:hand_human_lstm}}(a) and \figurename{~\ref{fig:hand_human_lstm}}(b), respectively.
These joints are physically connected in a tree-like structure (solid lines in \figurename{~\ref{fig:hand_human_lstm}}).
Since there are often correlation patterns among the ``symmetrical'' joints,
which can be useful for 3D pose estimation,
we also introduce direct links between them (dashed lines in \figurename{~\ref{fig:hand_human_lstm}}).
Therefore, the full dependency graph can be constructed for the human body as illustrated in \figurename{~\ref{fig:hand_human_lstm}}(a) and hand in \figurename{~\ref{fig:hand_human_lstm}}(b).

\textbf{Graphical ConvLSTM.}
As a successful extension of the recurrent neural networks,
long short-term memory (LSTM) networks \cite{hochreiter1997long}
can learn the complex long-term and short-term context dependency relations over the sequential input data.
Due to the natural dependencies among different parts of the human hand (or body),
LSTM is highly suitable for modeling the direct (``short-term'') and indirect (``longer-term'') dependency patterns among different parts for 3D pose estimation.

Since we aim to investigate the long short-term dependencies for boosting the feature maps within the CNN framework,
we adopt the convolutional LSTM (ConvLSTM) \cite{xingjian2015convolutional},
a variant of the original LSTM that can handle 2D input data,
as the main building block of our LSTD module.
Therefore, the inputs and outputs of the LSTD module are both feature maps.

Specifically,
to model the graphical dependencies among different parts,
instead of linking the units of the ConvLSTM sequentially as in \cite{xingjian2015convolutional},
we arrange and link the units of the ConvLSTM (see \figurename{~\ref{Figure:Network_Architecture}})
in our LSTD module by following the above-mentioned dependency graph
(see \figurename{~\ref{fig:hand_human_lstm}}).
We call this ConvLSTM design ``Graphical ConvLSTM''.

With the designed Graphical ConvLSTM,
the graphical long short-term dependency and context information is modeled unit by unit
in the LSTD module via the dependency links.
Moreover,
the correlations inside the dependency graph can be explored in two directions:
the forward pass (denoted as red arrows in \figurename{~\ref{fig:hand_human_lstm}})
and the backward pass (denoted as green arrows).
Thus, we can implement the Graphical ConvLSTM in a bidirectional fashion to allow the context information propagating in both directions inside the graph,
similar to the Bidirectional LSTM in \cite{graves2013speech}.



\textbf{Feature Boosting.}
Let $\mathds{F}$ denote the feature maps (channels) learned by the previous CNN layers,
as illustrated in \figurename{~\ref{Figure:Network_Architecture}}.
We equally divide $\mathds{F}$ to $J$ feature map groups,
\ie, $\mathds{F}=\{F_{1}, F_{2}, ..., F_{J}\}$,
where $F_{j}$ is the set of feature maps for representing the joint $j$ ($j \in [1, J]$),
and $J$ is the number of parts (joints).
Note that if $\mathds{F}$ cannot be divided equally, a $1\times1$ convolution can be performed on $\mathds{F}$ first.

Rather than directly feeding the feature maps (channels) $\mathds{F}$ to the subsequent convolutional layers to estimate the location of each joint, 
we boost them by feeding them to the LSTD module,
as depicted in \figurename{~\ref{Figure:Network_Architecture}}.
Concretely,
we feed the feature maps of each joint ($F_{j}$) to the corresponding unit ($j$) of the Graphical ConvLSTM as input,
and then the activations in this unit ($j$) are calculated as:
\begin{eqnarray}
\bar{\mathcal{H}}_j &=& \frac{1}{ |N_j| } \sum\limits_{k \in N_j} \mathcal{H}_{k} \label{eq:h_avg} \\
i_j &=& \sigma \big(F_{j} * W_{Fi} + \bar{\mathcal{H}}_j * W_{Hi} + b_i \big) \\
f_j &=& \sigma \big(F_{j} * W_{Ff} + \bar{\mathcal{H}}_j * W_{Hf} + b_f \big) \\
\tilde{c}_j &=& \tanh  \big(F_{j} * W_{Fc} + \bar{\mathcal{H}}_j * W_{Hc} + b_c \big) \\
\mathcal{C}_j &=& f_j \circ \big( \frac{1}{ |N_j| } \sum\limits_{k \in N_j} \mathcal{C}_{k} \big) + i_j \circ \tilde{c}_j  \label{eq:cj}\\
o_j &=& \sigma \big(F_{j} * W_{Fo} + \bar{\mathcal{H}}_j * W_{Ho} + b_o \big) \\
\mathcal{H}_{j} &=& o_{j}  \circ \tanh( \mathcal{C}_{j}) \label{eq:ht}
\end{eqnarray}
where $*$ denotes the convolution operator,
and $\circ$ indicates the Hadamard product.
$i_j$, $f_j$, $o_j$, and $\tilde{c}_j$ are respectively the input gate, forget gate, output gate, and modulated input for the unit $j$ in the ConvLSTM.
$\mathcal{C}_j$ and $\mathcal{H}_{j}$ are respectively the internal memory cell state and output hidden state of the unit $j$.
$N_j$ is the set of units linked to the unit $j$.

In our Graphical ConvLSTM,
each unit ($j$) may have links from more than one unit (\ie, $|N_j| > 1$).
For instance, joints 9 and 13 are both linked to joint 14 in the forward pass in \figurename{~\ref{fig:hand_human_lstm}(a)}.
Therefore, we aggregate the states of these linked units for current unit $j$ to represent its context information.
As formulated in Eq (\ref{eq:h_avg}),
average pooling is used for this aggregation operation
to obtain a fixed dimension of the context representation and avoid bringing extra parameters.

By using the aforementioned transition equations (\ref{eq:h_avg})\--(\ref{eq:ht}) at each unit ($j$) of the Graphical ConvLSTM,
we can then obtain the boosted feature maps ($\mathcal{H}_{j}$) for the joint $j$. 
By incorporating the context information of the graphical long short-term dependency with other parts,
into the input feature maps ($F_{j}$),
the produced feature maps ($\mathcal{H}_{j}$) from each unit of the Graphical ConvLSTM have more representational power for 3D pose estimation. 

\subsection{Context Consistency Gate}
\label{sec:method:gate}

In our network, 
the inputs at each unit ($j$) of the LSTD module are the feature maps ($F_{j}$) for representing a hand (or body) part.
However, the feature maps that are learned from a single RGB image by the previous convolutional layers
may be unreliable for representing the 3D pose information,
since there are often high ambiguities, heavy occlusions, and texture variations in the 3D pose estimation task.
The unreliable input feature maps can limit the performance of the feature boosting in the LSTD module,
and their propagation in the network may also affect the capability of the overall framework for 3D pose estimation.
In order to deal with the unreliable features,
in this paper,
we introduce a gating mechanism based on the LSTD module.
It assesses the consistency of the convolutional features to their context information,
accordingly adjusts them for feature boosting,
and regulates their propagation throughout the network.


The design of the gating function is inspired by the articulated nature of the human body's structure
and context consistency among the convolutional features 
for representing different hand (or body) parts.
Human joints are physically connected,
and the correlated joints form complex yet finite, common, and learnable patterns.
This indicates the state of a hand (or body) part is often consistent
with the context information of the whole hand (or body) structure. 
As a result, the feature maps extracted over an image for representing a hand (or body) part is supposed to be predictable
by using the context representations from other parts that are learned from the same image.

In previous works,
such as 
machine translation \cite{bahdanau2014neural},
video analysis \cite{liu2016eccv},
and image caption generation \cite{vinyals2015show},
LSTM networks that can model the dependency relations over the inputs
have demonstrated their ability in predicting the next input
based on the available context representations.
Inspired by the prediction ability of LSTM \cite{liu2016eccv}, 
we predict the input feature maps ($F_j$) at each unit $j$ of the Graphical ConvLSTM,
by using the available local context representations ($\{\mathcal{H}_{k}\}_{k \in N_j}$) from the linked units
and the global context information $\mathds{F}$ representing the entire human hand (or body). 
Concretely, we let the network learn a prediction of the input features at each unit ($j$)
as follows:
\begin{equation}
\mathcal{P}_j = \tanh \Big( \frac{1}{ |N_j| } \sum\limits_{k \in N_j} \big( \mathcal{H}_{k} * W_{Hp}^j \big) + \mathds{F} * W_{Fp}^j + b_p^j \Big)
\label{eq:Pj}
\end{equation}

We then assess the consistency of the input features at each unit ($j$) to the context representations
by
comparing the difference between the context-based prediction ($\mathcal{P}_j$) and the actual input feature maps ($F_j$).
Specifically,
we introduce a gating mechanism, 
context consistency gate (CCG), $\mathcal{G}_j$, to measure the consistency degree at the unit $j$ as:
\begin{equation}
\mathcal{G}_j = \exp \bigg( - \frac{\big( \mathcal{P}_j - \tanh(F_j) \big)^2}{{\omega}^2}  \bigg)
\label{eq:Gj}
\end{equation}
where $\omega$ is the weight to control the spread of the Gaussian function.
The outputs of this function vary between 0 and 1.


We then add the CCG to the designed Graphical ConvLSTM by modifying its cell state updating function (see Eq (\ref{eq:cj})) as:
\begin{equation}
\mathcal{C}_j = \Big( f_j \circ \big( \frac{1}{ |N_j| } \sum\limits_{k \in N_j} \mathcal{C}_{k} \big) \Big) \circ (1-\mathcal{G}_j)
               + \Big( i_j \circ \tilde{c}_j  \Big) \circ \mathcal{G}_j
\label{eq:NEWcj}
\end{equation}
The new cell state updating mechanism can be analyzed as follows.
If the input feature maps ($F_j$) are reliable (consistent with the context representations),
$\mathcal{G}_j$ is close to 1, and our LSTD module will import more information from them. 
Otherwise, if the feature maps are not consistent with the context,
\ie, $\mathcal{G}_j$ is close to 0,
the propagation of these feature maps is suppressed,
and the boosted features will be produced by exploiting more context information.

The CCG acts as a soft modulator to regulate the intermediate feature maps within the CNN architecture,
based on the estimation of the context consistency.
Therefore,
by adding CCG, our proposed LSTD module has more strength to boost the features for 3D pose estimation.

\subsection{Details of Network Structure}
\label{sec:method:details}

\textbf{Convolutional Layers.}
In our feature boosting network,
the state-of-the-art Hourglass CNN \cite{newell2016stacked} is adopted to learn the convolutional features for the RGB image,
as illustrated in \figurename{~\ref{Figure:Network_Architecture}}.
We follow the implementation in \cite{newell2016stacked} to construct each Hourglass CNN module,
such that the size of $\mathds{F}$ is $64 \times 64 \times 256$,
where $256$ is the number of feature maps (channels).


\textbf{LSTD Module with CCG.}
In the LSTD module,
the input size and cell state size at each unit are both $64 \times 64 \times 16$.
This indicates the number of feature maps for representing each joint is 16.
Since the bidirectional design is used for our Graphical ConvLSTM,
the output at each unit is calculated with a summation of the hidden state from the forward pass ($\overrightarrow{\mathcal{H}_{j}}$)
and the hidden state from the backward pass ($\overleftarrow{\mathcal{H}_{j}}$),
\ie, the boosted feature maps at unit $j$ are $\overrightarrow{\mathcal{H}_{j}}+\overleftarrow{\mathcal{H}_{j}}$.

\textbf{2D Heatmap Generation.}
We follow the recent works \cite{zhou2017towards,zimmermann2017learning} with a pipeline design,
which estimates the 2D heatmaps \cite{newell2016stacked} first as an intermediate representation for inferring the final 3D pose.
As shown in \figurename{~\ref{Figure:Network_Architecture}},
the boosted feature maps output from each unit ($j$) of the LSTD module are fed to the subsequent convolutional layers to generate the 2D heatmap ($H_j$) for the corresponding joint $j$.
The size of each heatmap ($H_j$) is $64 \times 64 \times 1$.
Readers are referred to \cite{newell2016stacked} for more details about the mechanism of 2D heatmaps.

\textbf{Depth Regression.}
We aggregate the generated 2D heatmaps and the boosted feature maps with a $1 \times 1$ convolution followed by a summation operation,
as shown in \figurename{~\ref{Figure:Network_Architecture}}.
Here the $1 \times 1$ convolution is used to map the 2D heatmap representations and boosted feature maps to the same size to facilitate the summation.
The aggregated representation with size $64 \times 64 \times 256$ is finally fed to a depth regression module
that contains four sequential convolutional layers with pooling and a fully connected layer for regressing the depth values of the joints.
Note that skip connection from the first layer of the Hourglass CNN is introduced to the aggregation operation.
This skip connection speeds up the convergence and enables the training of much deeper models, as analyzed by \cite{he2016deep}. 

After obtaining the depth value ($d_j$) of each joint ($j$), we can then combine the 2D representation ($H_j$) and the depth value ($d_j$) to produce the final 3D pose.

\textbf{Network Stacking.}
As illustrated in \figurename{~\ref{Figure:Network_Architecture}},
we stack multiple similar sub-networks to improve the representation capability of our network for 3D pose estimation.
Each sub-network contains the Hourglass CNN layers for feature learning
and an LSTD module with CCG for feature boosting.
With this stacking design, the convolutional features are boosted at multiple levels within the network.
Two sub-networks are stacked in our implementation.

\textbf{Objective Function.}
The objective function of our network is formulated as:
$\ell=  \ell_{H} + \gamma \ell_{D}$,
where $ \ell_{H}$ is the mean squared error
measuring the difference between the ground truth 2D heatmaps ($H_{j}$) and the prediction results ($\hat{H}_{j}$).
$\ell_{D}$  is the mean squared error
measuring the difference between the ground truth depth values ($d_{j}$) and the regressed values ($\hat{d}_{j}$).
The whole network is trained in an end-to-end fashion by stochastic gradient descent optimization.

\section{Experiments}
\label{sec:experiments}

The proposed approach is evaluated on
the 3DHandPose dataset \cite{zhang20163d} for hand pose estimation,
and the Human3.6M dataset \cite{ionescu2014human3} for body pose estimation.
The MPI-INF-3DHP \cite{mehta2017monocular} and MPII \cite{andriluka20142d} datasets are also used for qualitative analysis.
We conduct extensive experiments using the following models to test our proposed method:
\\
  (1) 3D Pose Net.
  This is the baseline network model for 3D pose estimation.
  In this network, the CNN feature maps without feature boosting are fed to the CNN layers for 2D heatmap generation and depth regression.
  \\
  (2) 3D Pose Net with FB.
  In this network, the LSTD module proposed by us is used for feature boosting (FB).
  However, the CCG is not added.
  \\
  (3) 3D Pose Net with FB+. This is the proposed feature boosting network for 3D pose estimation.
  The LSTD module is embedded in the CNN framework for feature boosting,
  and the CCG is also added to improve the design of the LSTD module.

\subsection{Implementation Details}
\label{sec:experiments:details}

In our experiment,
the parameter $\gamma$ in the objective function is set to 0.1,
and ${\omega}^2$ in Eq (\ref{eq:Gj}) is set to 2.
These hyper-parameters are obtained by using cross-validation protocol on the training sets,
and the parameter set achieving the optimum performance is used.
The hourglass CNN layers are implemented by following \cite{newell2016stacked}.
%
Data augmentation is used in our experiments, including random translation, scaling, and rotation.

During training, the 3D pose is aligned to the 2D pose of the image plane,
\ie, aligning the root joint location and also the human body scale.
Then this aligned 3D pose (at the image pixel level) is used for network training.
In testing, the estimated 3D pose is re-scaled to the size of a pre-defined canonical skeleton, as done in \cite{zhou2018monocap,zhou2017towards}.
Rigid transformation \cite{nie2017monocular} is not used in our experiment.
For evaluation, the estimated pose and ground truth pose are aligned based on the root joint locations.

\subsection{Experiments of 3D Hand Pose Estimation}
\label{sec:experiments:handpose}

The 3DHandPose dataset \cite{zhang20163d} is a large dataset for 3D hand pose estimation.
It is captured under varying illumination conditions with 6 different backgrounds.
Different from the NYU Hand Pose dataset \cite{tompson2014real},
which is mainly designed for hand pose estimation from depth images and the registered color images contain lots of artifacts,
the large 3DHandPose dataset is highly suitable for 3D hand pose estimation from a single RGB image, as analyzed in \cite{zimmermann2017learning}.
In this dataset, the 2D and 3D annotations of 21 keypoints of the human hand are provided for each frame.
We follow the evaluation protocol of \cite{zimmermann2017learning} by using 30,000 hand images for training and 6,000 hand images for testing.

\begin{figure}[!tb]
	\centerline{\includegraphics[scale=0.56,trim={35 305 135 270},clip]{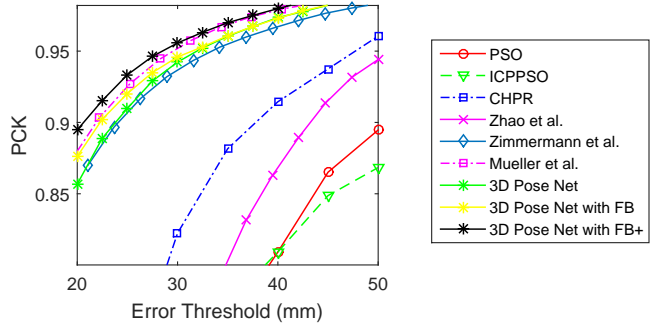}}
	\caption{3D hand pose estimation results on the 3DHandPose dataset.
The curves indicate the percentage of correct keypoint (PCK) over the respective threshold in \emph{mm}.}
	\label{fig:resultHandCurve}
\end{figure}

The experimental results are shown in \figurename{~\ref{fig:resultHandCurve}} and \tablename{~\ref{table:result3DPCKHand}}.
We report the percentage of correct keypoints (PCK) for different error thresholds on this dataset by following \cite{zimmermann2017learning}.
The results show that our proposed method outperforms the other methods on this dataset.

The 3D Pose Net and the model proposed by Zimmermann \etal \cite{zimmermann2017learning} are both CNN-based methods
without considering the dependency structure of the features of the hand joints,
thus their performances are inferior to the proposed feature boosting network with the LSTD module.
By adding the CCG to the LSTD module,
the performance of our method (3D Pose Net with FB+) is further improved.

\begin{figure}[!tbp]
	\centerline{\includegraphics[scale=0.23,clip]{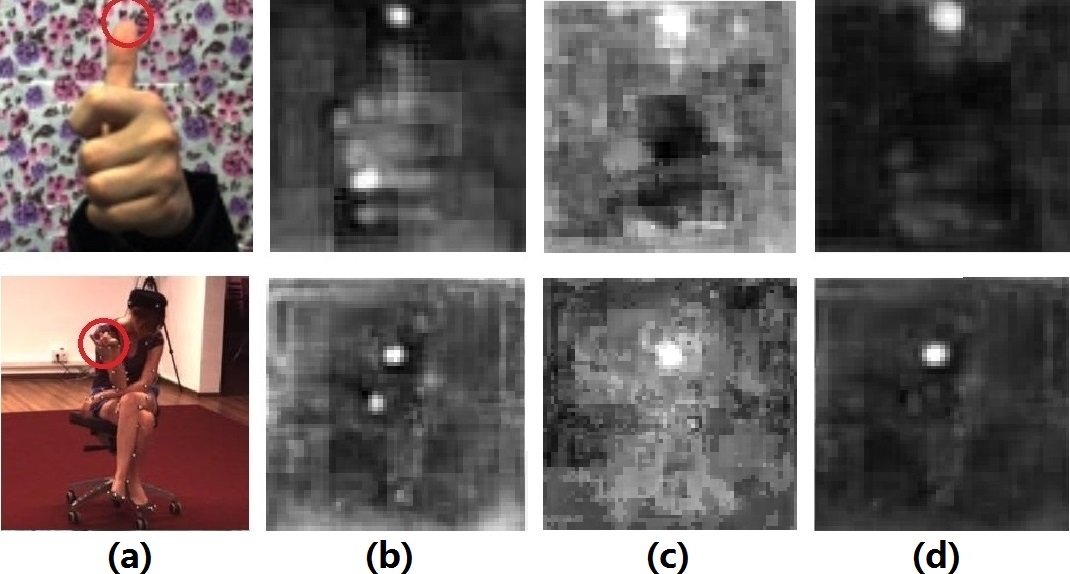}}
	\caption{Visualization of feature maps before and after boosting for different joints (labeled as red circles).
The four columns are respectively (a) input image, (b) feature map for representing a joint before boosting, (c) CCG, and (d) feature map after boosting.
}
	\label{fig:visualizeFM}
\end{figure}

\begin{table}[!tb]
	\caption{Experimental results on the 3DHandPose dataset.
Numbers are percentage of correct keypoint (PCK) over respective threshold in \emph{mm}.
Refer to \figurename{~\ref{fig:resultHandCurve}} for more results.}
	\label{table:result3DPCKHand}
	\centering
	\scriptsize
	\begin{tabular}{|c|ccc|}
        \hline
		Error Threshold (\emph{mm})                          & PCK@20              &  PCK@25         & PCK@30   \\
		\hline
        PSO \cite{zhang20163d}                               &    32.2\%             &   54.0\%          &  67.4\%         \\
        Zhao \etal \cite{zhao2017simple}                     &    43.6\%             &   56.8\%          &  70.1\%  \\
        ICPPSO \cite{zhang20163d}                            &    52.0\%             &   64.5\%          &  71.7\%         \\
        CHPR \cite{zhang20163d}                              &    56.6\%             &   71.7\%          &  82.2\%         \\
        Zimmermann \etal \cite{zimmermann2017learning}       &    85.9\%             &   90.7\%          &  93.7\%     \\
        Mueller \etal \cite{mueller2018ganerated}            &    88.0\%             &   92.5\%          &  95.2\%     \\
        \hline 
        3D Pose Net                                          &    85.7\%             &   91.0\%          &  94.2\%     \\
        3D Pose Net with FB                                  &    87.7\%             &   92.1\%          &  94.6\%     \\
        3D Pose Net with FB+                                 &   \textbf{89.5\%}     &  \textbf{93.3\%}  &  \textbf{95.6\%}  \\
		\hline
	\end{tabular}
\end{table}


Since the graphical long short-term dependency relations among the joints are modeled in our network,
we also evaluate the performance of the network by using different dependency connections, and report the results in \tablename{~\ref{table:result3DConvLSTM}}.
The ``Simple Sequence'' means that the hand joints are linked one by one as a sequential chain by following the enumeration order.
The ``Physical Dependency'' link indicates that the real physical connections between the joints are used (as shown by the solid lines in \figurename{~\ref{fig:hand_human_lstm}}).
The ``Symmetrical Connections'' means that the ``symmetrical'' relations are used (dashed lines in \figurename{~\ref{fig:hand_human_lstm}}).
The ``Graphical Dependency'' link indicates that both the physical and ``symmetrical'' connections are used, but only the forward pass is enabled.
The ``Bi-directional Graphical Dependency'' is the proposed graphical long short-term dependency relationship with bidirectional passes, as shown in \figurename{~\ref{fig:hand_human_lstm}}.
The results in \tablename{~\ref{table:result3DConvLSTM}} show that the ``Graphical Dependency'' is superior to the ``Physical Dependency'' only and the ``Symmetrical Connections'' only,
which indicates that it is beneficial to combine the ``symmetrical'' relation links and the physical dependency links for pose estimation.
Our proposed ``Bi-directional Graphical Dependency'' yields the best result for 3D hand pose estimation, as shown in \tablename{~\ref{table:result3DConvLSTM}}.

\begin{table}[!b]
	\caption{Evaluation of using different connections for ConvLSTM on the 3DHandPose dataset.}
	\label{table:result3DConvLSTM}
	\centering
	\scriptsize
	\begin{tabular}{|c|c|}
		\hline
        Connections & Accuracy (PCK@20) \\
        \hline
        Simple Sequence & 86.1\% \\
        Physical Dependency & 87.5\%\\
        ``Symmetrical'' Connections & 87.4\% \\
        Graphical Dependency & 89.0\%\\
        Bi-directional Graphical Dependency & 89.5\%\\
        \hline
	\end{tabular}
\end{table}

We evaluate the performance of the proposed framework with different numbers of the sub-networks for feature learning and boosting, and show the results in \tablename{~\ref{table:resultStacks}}.
The results show that our feature boosting network with two sub-networks outperforms the single sub-network framework.
This indicates that by boosting the feature maps at multiple levels, the 3D pose estimation performance can be improved.
Due to the memory limitation of our GPUs, we were not able to try stacking more sub-networks.
\begin{table}[!b]
	\caption{Evaluation of the feature boosting network with different numbers of sub-networks.}
	\label{table:resultStacks}
	\centering
	\scriptsize
	\begin{tabular}{|c|c|}
		\hline
        ~~~~Network stacking~~~~ & ~~Accuracy (PCK@20)~~ \\
        \hline
        One sub-network  & 87.4\% \\
        Two sub-networks & 89.5\%\\
        \hline
	\end{tabular}
\end{table}

We also visualize some examples of the feature maps in our network,
as illustrated in \figurename{~\ref{fig:visualizeFM}}.
Specifically, we visualize the feature maps learned by the previous CNN layers before feature boosting,
and the boosted feature maps.
The results show that by using the LSTD module with CCG for boosting,
the produced feature maps are more reliable and stable compared to the feature maps before boosting.

\begin{table*}[!tbp]
	\caption{Comparison with the state-of-the-art work on 3D body pose estimation on the Human3.6M dataset.
Numbers are the mean Euclidian distance (\emph{mm}) between the estimated 3D joints and the ground truth joints.}
	\label{table:result3DHuman36M}
	\centering
	\scriptsize
	\begin{tabular}{|c|p{0.5cm}p{0.5cm}p{0.5cm}p{0.5cm}p{0.5cm}p{0.5cm}p{0.5cm}p{0.69cm}p{0.5cm}p{0.65cm}p{0.5cm}p{0.5cm}p{0.69cm}p{0.5cm}p{0.72cm}p{0.72cm}|}
\hline
		\textbf{Method}                         & \textbf{Direct} & \textbf{Discuss}  &   \textbf{Eat}    &  \textbf{Greet}  &  \textbf{Phone}  &   \textbf{Photo}  &   \textbf{Pose}   &  \textbf{Purchase}  & \textbf{Sit}       & \textbf{SitDown}  & \textbf{Smoke}    & \textbf{Wait}  & \textbf{WalkDog} & \textbf{Walk}  & \textbf{WalkPair} &\textbf{Average}\\
		\hline 
        Tome \etal \cite{tome2017lifting}       &   64.98            &   73.47           & 76.82             & 86.43            & 86.28            & 110.67            & 68.93             & 74.79               & 110.19             &173.91             &84.95              &85.78             &86.26             &71.36             & 73.14             &88.39  \\
        Metha \etal \cite{mehta2017monocular}   & 57.51              &   68.58           & 59.56             & 67.34            & 78.06            & 82.40             & 56.86             & 69.13               & 99.98              &  117.53           &   69.44           & 67.96            & 76.50            & 55.24            &61.40              & 72.88 \\
        Pavlakos \etal \cite{pavlakos2017coarse}&    67.38           &   71.95           &  66.70            & 69.07            & 71.95            & 76.97             & 65.03             & 68.30               &   83.66            & \textbf{96.51}    & 71.74             &65.83             &74.89             &59.11             &63.24              &71.90 \\
        Nie \etal \cite{nie2017monocular}       &     90.10           &  88.20           &        85.70      &     95.60        &       103.90     &  103.00           & 92.40             &  90.40              & 117.90             &  136.40           &    98.50          &   94.40          &    90.60         &  86.00           &  89.50            & 97.50   \\
		Zhou \etal \cite{zhou2018monocap}       &   71.40            &   77.00           &  75.70            &    77.20         &  76.60           &     102.30        &       79.30       &     75.00           &     76.00          &    112.20         &   74.20           &     91.30        &      73.10       &  57.80           &    74.10          & 79.60 \\
        Rhodin \etal \cite{rhodin2018learning} & - & -& -& -& -& -& -&- &- & -&- &- & -& -&- & 66.80 \\
        Zhou \etal \cite{zhou2017towards}       &     54.82          &    60.70          &  58.22            & 71.41            &  62.03           &  \textbf{65.53}   & 53.83             & 55.58               & 75.20              & 111.59            & 64.15             & 66.05            &\textbf{51.43}    &63.22             &55.33              & 64.90  \\
        \hline 
        Proposed                                &  \textbf{50.72}    &  \textbf{60.04}   &  \textbf{51.11}   &  \textbf{63.65}  &  \textbf{59.70}   &  69.34            &   \textbf{48.83}   & \textbf{51.98}  &  \textbf{72.76}    &   105.31          & \textbf{58.62}    &  \textbf{60.98}  & 62.25            &  \textbf{45.88}  & \textbf{48.69}   &\textbf{61.10}\\ \hline
	\end{tabular}
\end{table*}

\subsection{Experiments of 3D Body Pose Estimation}
\label{sec:experiments:human36m}

\textbf{Human3.6M.}
The Human3.6M dataset \cite{ionescu2014human3} is a large-scale and widely used dataset for 3D human body pose estimation.
This dataset contains 3.6 million human poses captured with a motion capture system.
%
We follow the evaluation protocol in \cite{zhou2017towards} on this dataset,
in which 5 subjects (s1, s5, s6, s7, and s8) are used for training,
and 2 subjects (s9 and s11) are adopted for testing.
The videos in this dataset are down-sampled from $50fps$ to $10fps$.
The training sample combination in \cite{zhou2017towards} is adopted to train our network
(half Human3.6M data \cite{ionescu2014human3} and half MPII data \cite{andriluka20142d}).

The experimental results (PCKs) on the Human3.6M dataset are shown in \tablename{~\ref{table:result3DPCKHuman36M}}.
The results show that by using the LSTD module with CCG for feature boosting, the ``3D Pose Net with FB+'' achieves the best results.
We also compare the proposed feature boosting network with the state-of-the-arts, and report the results
in \tablename{~\ref{table:result3DHuman36M}}. 
We can observe that the feature boosting network outperforms other methods for 3D human pose estimation.

\begin{table}[!tbp]
	\caption{Experimental results on the Human3.6M dataset.
             }
	\label{table:result3DPCKHuman36M}
	\centering
	\scriptsize
	\begin{tabular}{|c|ccc|c|}
        \hline
		Error Threshold (\emph{mm})                    &  PCK@50            &  PCK@75            & PCK@100             & Mean Error    \\
		\hline
        3D Pose Net                                    &         48.1\%     &          68.9\%    &        80.1\%       &     65 \emph{mm}      \\
        3D Pose Net with FB                            &         49.8\%     &         69.9\%     &        81.5\%       &   63 \emph{mm}    \\
        3D Pose Net with FB+                           &  \textbf{51.5\%}   &  \textbf{71.4\%}   &     \textbf{82.8\%} & \textbf{61 \emph{mm}} \\
		\hline
	\end{tabular}
\end{table}


We also follow the data processing and evaluation setting of \cite{sun2017compositional},
and use the videos of 5 subjects for training, while evaluating on 2 subjects by using 1 frame from every 64 frames.
On this setting, the joint error of our method is 58.0 \emph{mm},
which is lower than 59.1 \emph{mm} of the method in \cite{sun2017compositional}.

\textbf{Cross-dataset evaluation on MPI-INF-3DHP.}
We perform cross-dataset evaluation on the MPI-INF-3DHP \cite{mehta2017monocular} dataset,
\ie, only Human3.6M and MPII are used for training, while the testing is performed on MPI-INF-3DHP.
We follow the evaluation criteria in \cite{zhou2017towards} and report the average PCK in \tablename{~\ref{table:result3DPCKMPIINF3DHP}}.
The results show that our proposed feature boosting network achieves good performance in this cross-dataset evaluation scenario.

\begin{table}[!tbp]
	\caption{Experimental results on MPI-INF-3DHP. }
	\label{table:result3DPCKMPIINF3DHP}
	\centering
	\scriptsize
	\begin{tabular}{|c|cc|cc|}
        \hline
		Method        & ~~~\cite{zhou2017towards}~~~ &   ~~~\cite{mehta2017monocular}~~~ &  3D Pose Net     & 3D Pose Net with FB+    \\
		\hline
        PCK           &    69.2\%              &  64.7\%                &  66.2\%          &        69.6\%          \\
		\hline
	\end{tabular}
\end{table}

\begin{figure}[!tbp]
	\centerline{\includegraphics[scale=0.375,trim={0cm  11.5cm 3.0cm 0.1cm},clip]{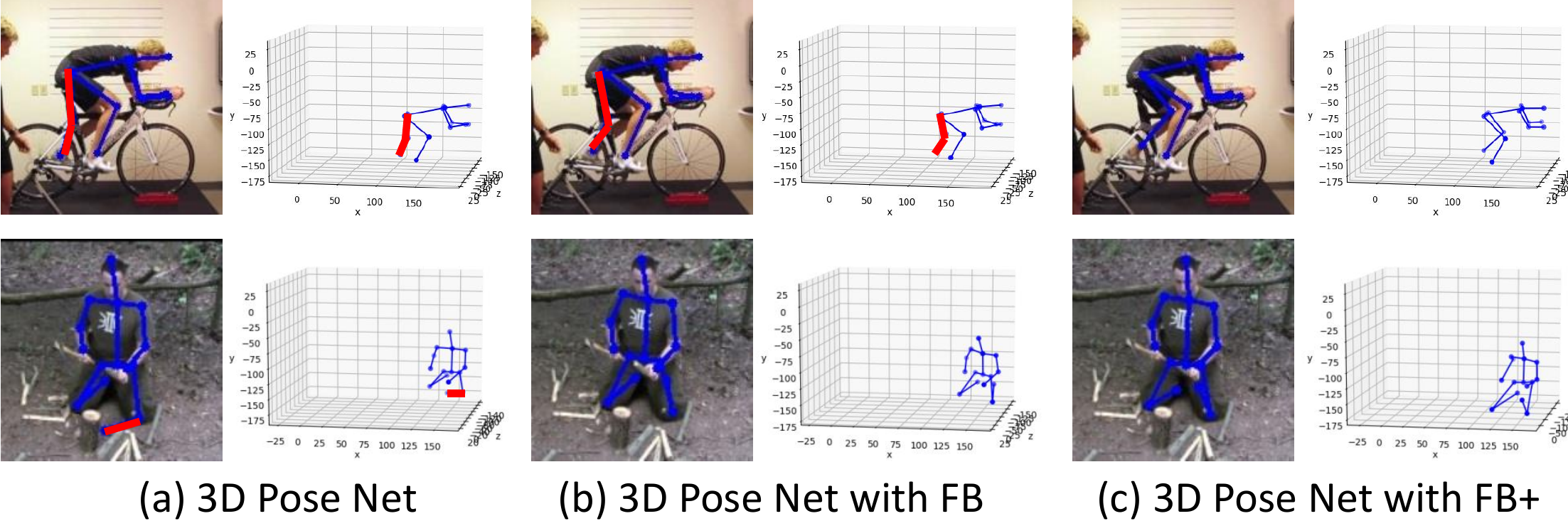}}
	\caption{Qualitative results on MPII. The wrongly estimated joints are depicted as red lines.
}
	\label{fig:ResultsCompare}
\end{figure}

\begin{figure}[!tbp]
	\centerline{\includegraphics[scale=0.46,trim={0.05cm 3.9cm 6.9cm 3.6cm},clip]{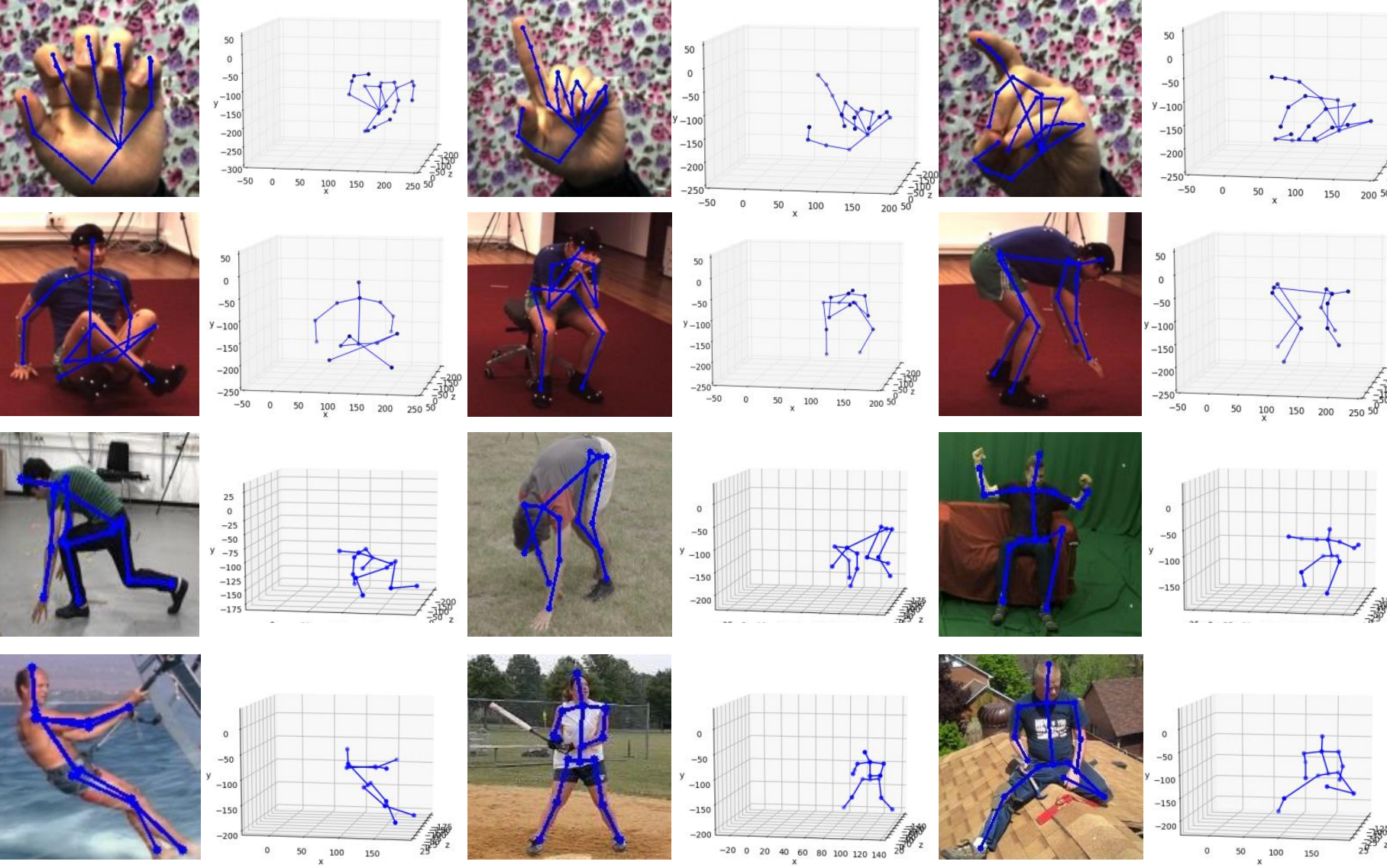}}
\caption{Results on 3DHandPose (top row), Human3.6M (2nd row), MPI-INF-3DHP (3rd row), and MPII (4th row).}
	\label{fig:Results}
\end{figure}

\textbf{Qualitative evaluation on MPII validation subset.}
The 3D pose annotations are not provided in MPII \cite{andriluka20142d} dataset,
and we use its validation subset for qualitative evaluation.
The network trained for Human3.6M is used for evaluation on the MPII validation subset.
We visualize some of the challenging examples in \figurename{~\ref{fig:ResultsCompare}} and \figurename{~\ref{fig:Results}}.
The results show that our proposed feature boosting network can reliably handle the challenging poses,
as shown in \figurename{~\ref{fig:ResultsCompare}(c)}.

\subsection{More Experiments}
\label{sec:experiments:moreExp}

\begin{figure}
    \begin{minipage}[b]{0.49\linewidth}
		\centering
		\centerline{\includegraphics[scale=0.17,trim={6.0cm 0.0cm 18.8cm 0.9cm},clip]{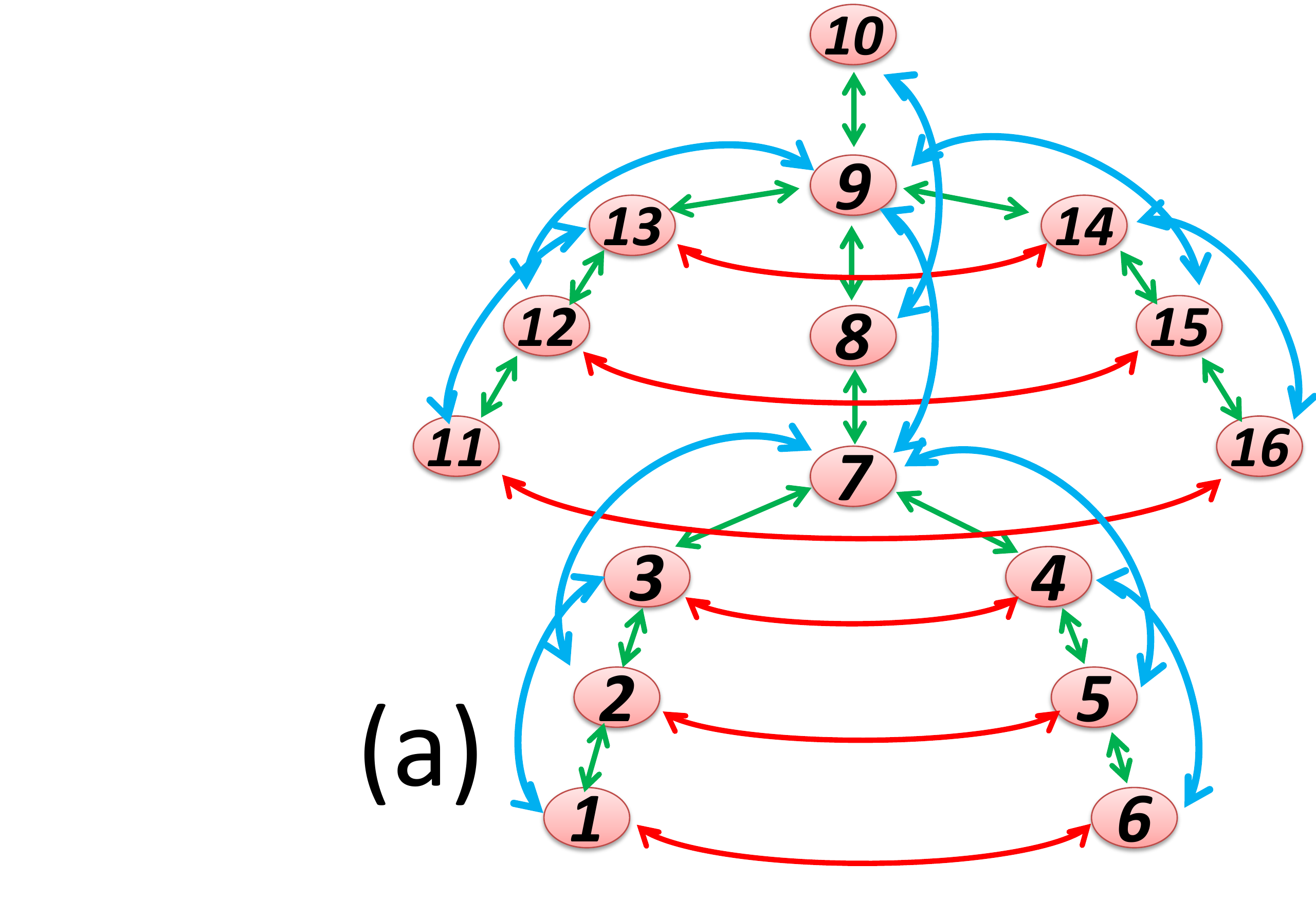}}
	\end{minipage}
	\begin{minipage}[b]{0.49\linewidth}
		\centering
		\centerline{\includegraphics[scale=0.2,trim={14.8cm 3.2cm 19.8cm 2.6cm},clip]{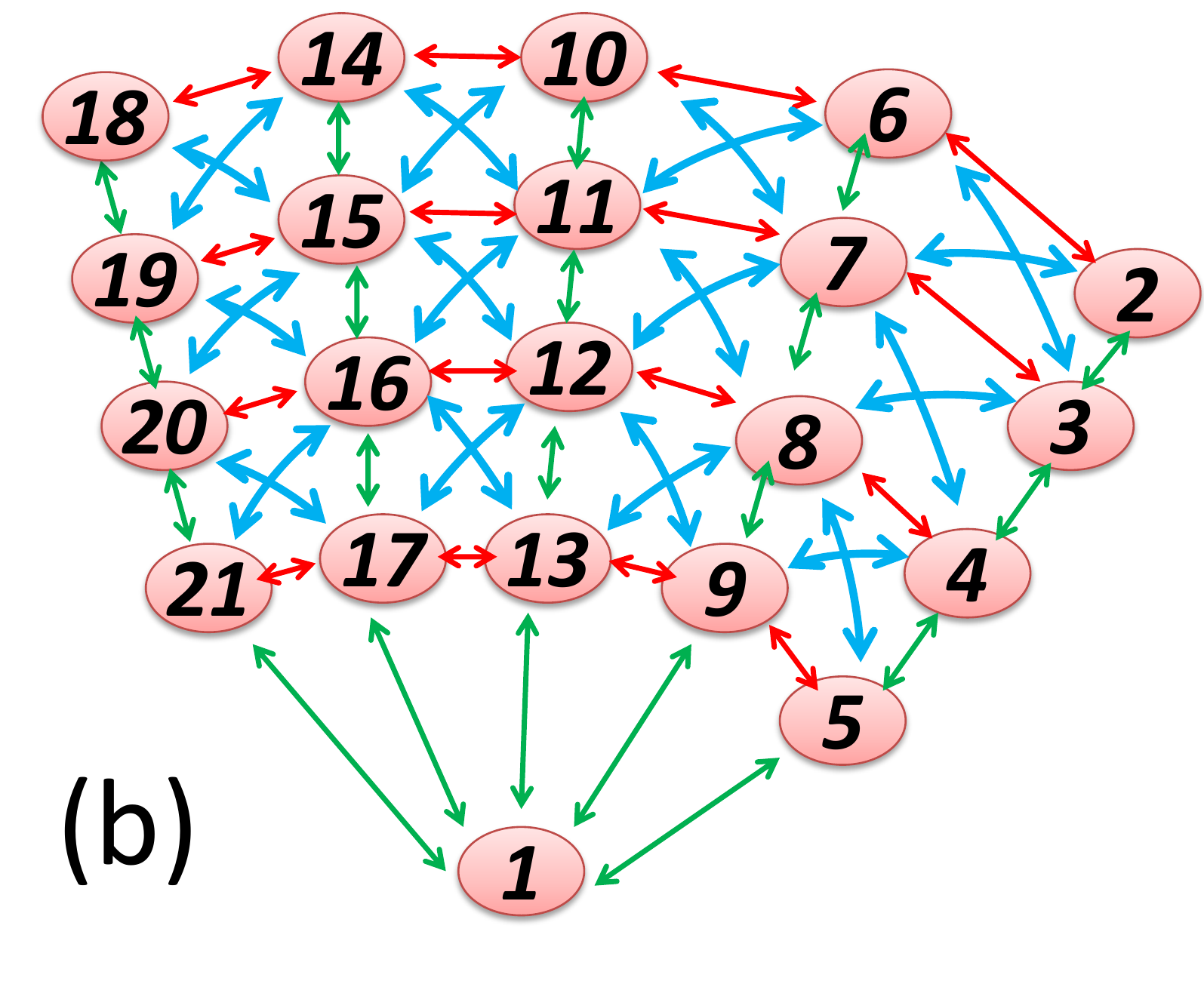}}
	\end{minipage}
	\caption{Illustration of involving more connections.
            Extra links (denoted as blue arrows) are added. 
            }
	\label{fig:hand_human_lstm_moreConn}
\end{figure}

\textbf{Evaluation of involving more connections.}
In \figurename{~\ref{fig:hand_human_lstm}}, except joint 1 of the hand, the maximum number of linked joints is 4.
Here we investigate the performance of our network by involving more connections.
We add extra links to the dependency graph,
such that the maximum numbers of linked joints for body and hand become 7 and 8, respectively,
as shown in \figurename{~\ref{fig:hand_human_lstm_moreConn}}.
The results in \tablename{~\ref{table:resultMoreConn}} show that when involving extra links,
the accuracy does not improve (on 3DHandPose) or only improves a little bit (on Human3.6M).
This also shows the effectiveness of our designed graphical long short-term dependency relationship in \figurename{~\ref{fig:hand_human_lstm}}.

\begin{table}[!tbp]
	\caption{Evaluation of involving more connections. }
	\label{table:resultMoreConn}
	\centering
	\scriptsize
	\begin{tabular}{|c|cc|cc|}
        \hline
        \multirow{2}{*}{Dataset}          &  \multicolumn{2}{c|}{Human3.6M}  &  \multicolumn{2}{c|}{3DHandPose} \\
                                          &  \multicolumn{2}{c|}{(PCK@50)}   &  \multicolumn{2}{c|}{(PCK@20)}   \\
        \hline
		Max. number of linked joints      &    4      &    7                 &   4               & 8            \\
        Accuracy  (\%)                    &    51.5\% &  51.7\%              &  89.5\%           & 89.5\%       \\
		\hline
	\end{tabular}
\end{table}

\textbf{Evaluation of using different recurrent models.}
Our LSTD module is designed based on the ConvLSTM structure.
We also evaluate the performance of our network by using different recurrent structures, namely ConvLSTM, ConvRNN, and ConvGRU,
and report the results in \tablename{~\ref{table:resultDifferentRNN}}.
The results show that the accuracy of ConvLSTM is higher than ConvRNN and ConvGRU.
We also observe that the 3D Pose Net with FB using different recurrent structures all outperforms 3D Pose Net. 

\begin{table}[!tbp]
	\caption{Evaluation of using different recurrent models. }
	\label{table:resultDifferentRNN}
	\centering
	\scriptsize
	\begin{tabular}{|c|c|c|}
        \hline
        \multirow{2}{*}{Dataset}              &  Human3.6M  &  3DHandPose \\
                                              &  (PCK@50)   & (PCK@20)   \\
        \hline
        3D Pose Net with FB (ConvRNN)         &    49.3\%   & 87.0\%       \\
        3D Pose Net with FB (ConvGRU)         &    49.7\%   & 87.5\%       \\
        3D Pose Net with FB (ConvLSTM)        &    49.8\%   & 87.7\%       \\
		\hline
	\end{tabular}
\end{table}

\textbf{2D pose.}
Since 2D pose is estimated in our network, we also evaluate its performance and report the results in \tablename{~\ref{table:result2DHandPose}}.
The standard PCK metric is used for evaluation, and the distance is normalized by the finger width (referred to as PCKf).
We observe that the 2D pose performance of the 3D Pose Net is lower than the Hourglass model (2-stacked).
This may be owing to the extra depth estimation task in 3D Pose Net.
Nevertheless, our proposed 3D Pose Net with FB+ yields a better result for 2D heatmap estimation than the Hourglass model.

\begin{table}[!tbp]
	\caption{2D pose accuracy on the 3DHandPose dataset. }
	\label{table:result2DHandPose}
	\centering
	\scriptsize
	\begin{tabular}{|c|c|cc|}
        \hline
		Method     & Stacked Hourglass   &   3D Pose Net    & 3D Pose Net with FB+     \\
		\hline
        PCKf@0.5   &   86.5\%            &  85.4\%          & 87.7\%          \\
		\hline
	\end{tabular}
\end{table}

\section{Conclusion}
\label{sec:conclusion}

We propose a feature boosting network for 3D hand and full body pose estimation in this paper.
A novel LSTD module is introduced to enable the convolutional features
to perceive the graphical long short-term dependency relationship among different hand (or body) parts.
The design of the LSTD module is further enhanced
by assessing the context consistency of the features with the CCG.
The proposed feature boosting network achieves state-of-the-art performance on challenging datasets for 3D hand and body pose estimation.

\end{document}